\documentclass[12pt]{article}
\usepackage[utf8]{inputenc}
\usepackage{geometry}
\usepackage{graphicx}
\usepackage[skip=0.1\baselineskip]{caption}
\usepackage{multirow}
\usepackage{comment}
\usepackage{url}
\usepackage{enumitem}
\setlist[enumerate]{itemsep=0mm}
\usepackage{titlesec}
\usepackage{gensymb}
\usepackage{booktabs} 
\usepackage{abstract}

\titleformat{\section}
  {\normalfont\fontsize{14}{15}\bfseries}{\thesection}{1em}{}
\titleformat{\subsection}
  {\normalfont\fontsize{12}{15}\bfseries}{\thesubsection}{1em}{}
\titleformat{\subsubsection}
  {\normalfont\fontsize{11}{15}\bfseries}{\thesubsubsection}{1em}{}
  
\titlespacing\section{2pt}{12pt plus 4pt minus 2pt}{0pt plus 2pt minus 2pt}
\titlespacing\subsection{0pt}{12pt plus 4pt minus 2pt}{0pt plus 2pt minus 2pt}
\titlespacing\subsubsection{0pt}{12pt plus 4pt minus 2pt}{0pt plus 2pt minus 2pt}

\newcommand{\at}{\makeatletter @\makeatother}

\usepackage{fancyhdr}
\title{\vspace{-1cm}\Large{Methods for Combining and Representing Non-Contextual Autonomy Scores for Unmanned Aerial Systems}}
\author{\small{Brendan Hertel, Ryan Donald, Christian Dumas, S. Reza Ahmadzadeh\thanks{Authors are with the Persistent Autonomy and Robot Learning (PeARL) lab at the University of Massachusetts Lowell, Lowell, MA, 01854.
Emails: \texttt{\{Brendan\_Hertel, Ryan\_Donald, Christian\_Dumas\}\at student.uml.edu, Reza\_Ahmadzadeh\at uml.edu.} \newline Address: 1 University Ave. Lowell, MA, 01854. P: 978-934-6175.}}}

\date{}

\begin{document}

\maketitle

\begin{abstract}
    \textbf{Abstract--}Measuring an overall autonomy score for a robotic system requires the combination of a set of relevant aspects and features of the system that might be measured in different units, qualitative, and/or discordant. In this paper, we build upon an existing non-contextual autonomy framework that measures and combines the \emph{Autonomy Level} and the \emph{Component Performance} of a system as overall autonomy score. We examine several methods of combining features, showing how some methods find different rankings of the same data, and we employ the weighted product method to resolve this issue. Furthermore, we introduce the non-contextual autonomy coordinate and represent the overall autonomy of a system with an autonomy distance. We apply our method to a set of seven  Unmanned Aerial Systems (UAS) and obtain their absolute autonomy score as well as their relative score with respect to the best system. 
\end{abstract}

\section{Introduction}

Autonomy is a concept which is continually aspired for in robotics. As robots become more present in the world, for many operations they must become more human-independent, or autonomous. Determining an actual autonomy score for a robotic system, however, has been counter intuitive and equivocal due to the presence of various factors measured in different units. Combining these factors in a meaningful way has proven to be difficult, and many existing frameworks for measuring autonomy are only applicable partially in special cases~\cite{clothier2014review}. In this paper, we build upon an existing framework and propose a novel method for determining the overall potential autonomy of a system.


A framework which can adequately measure autonomy must satisfy certain requirements. Firstly,  the autonomy score must be reproducible which means  measurements must be quantitative, not qualitative~\cite{gyagenda2017non}. Additionally, the autonomy score must have absolute meaning, not relative meaning~\cite{clothier2014review}. Scales which have relative meanings can only be used for  comparing the autonomy of systems, and do not convey any information about the absolute autonomy of the system. Finally, the autonomy score must be a meaningful combination of the relevant factors and properties of the system~\cite{gyagenda2017non}.

Existing autonomy evaluation methods can be divided into two main categories: contextual and non-contextual~\cite{durst2014levels}. Contextual autonomy evaluation methods take into account features from both the mission and the environment (e.g., ALFUS~\cite{huang2004autonomy}). Whereas non-contextual methods only rely on implicit system capabilities and do not consider the mission and environment features (e.g., NCAP~\cite{durst2011non}). Contextual autonomy evaluation methods provide mission-specific measures of the system based on the complexity of the mission and the environment where the operation takes place. Non-contextual autonomy evaluation methods, on the other hand, provide a potential measure of the system independence as an overall expected score.

In this paper, we focus on the non-contextual autonomy evaluation methods. We specifically employ and build upon the Non-Contextual Autonomy Potential, or NCAP~\cite{durst2011non}, for evaluating seven small Unmanned Aerial Systems (UAS) with different capabilities and specifications. We then discuss various methods for combining autonomy scores and their advantages and disadvantages.      

\section{Background on Non-Contextual Autonomy Potential (NCAP)}

NCAP divides the architecture of a UAS into four layers: perception, modeling, planning, and execution~\cite{durst2011non}. Each of these four layers builds off of the previous layer. The UAS uses its sensors to acquire raw data from the environment, encodes the data in a useful way to build a model (e.g., a map), uses its algorithms to compute a plan (i.e. path), and executes a plan in the real-world. These four layers form a loops starts from perception, goes through modeling, planning, and execution, then updates the states and repeats. While the information acquisition and execution steps depend on the quality and quantity of the system's hardware, the modeling and planning (and to some extent the execution) steps depend on the quality of the system's software (i.e., algorithms). 

In evaluating the perception layer, we consider both proprioceptive and exteroceptive sensors that are used to perceive the robot's status and its surroundings. The modeling layer includes algorithms that use the collected raw data to build an abstract model of the environment. Modeling includes various operations such as mapping, localization, target and obstacle detection. The planning layer is where the UAS uses the internally built model to plan a sequence of actions to reach the goal, which is considered a high-level human-provided knowledge such as safety concerns and mission goals. The planning layer might generate multiple plans but should be able to distinguish an optimal plan. Planning can include either path or behavior generation. The generated optimal plan includes a sequence of actions that should be executed by the UAS without requiring human assistance. 
Note that we consider robots within the same family (i.e., small unmanned aerial systems) and comparing robot autonomy for systems in different families is beyond the scope of this paper.

\section{UAS Platforms}
\label{platforms}

\begin{figure}[t]
\centering
\minipage{.14\textwidth}
	\includegraphics[width=\linewidth]{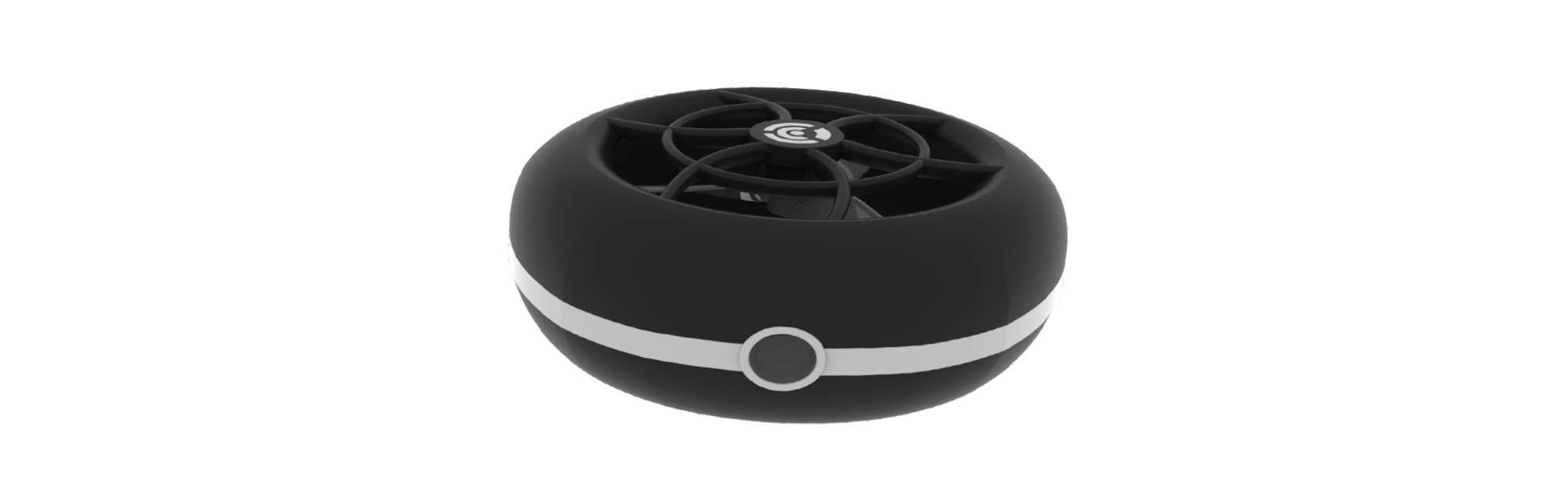}
\endminipage\hfill
\minipage{.14\textwidth}
	\includegraphics[width=\linewidth]{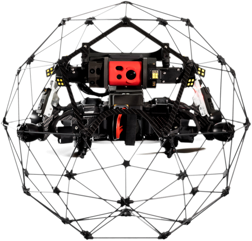}
\endminipage\hfill
\minipage{.14\textwidth}
	\includegraphics[width=\linewidth]{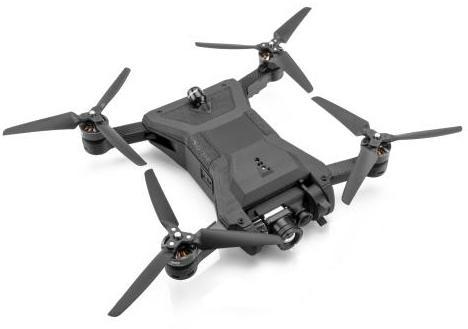}
\endminipage\hfill
\minipage{.14\textwidth}
	\includegraphics[width=\linewidth]{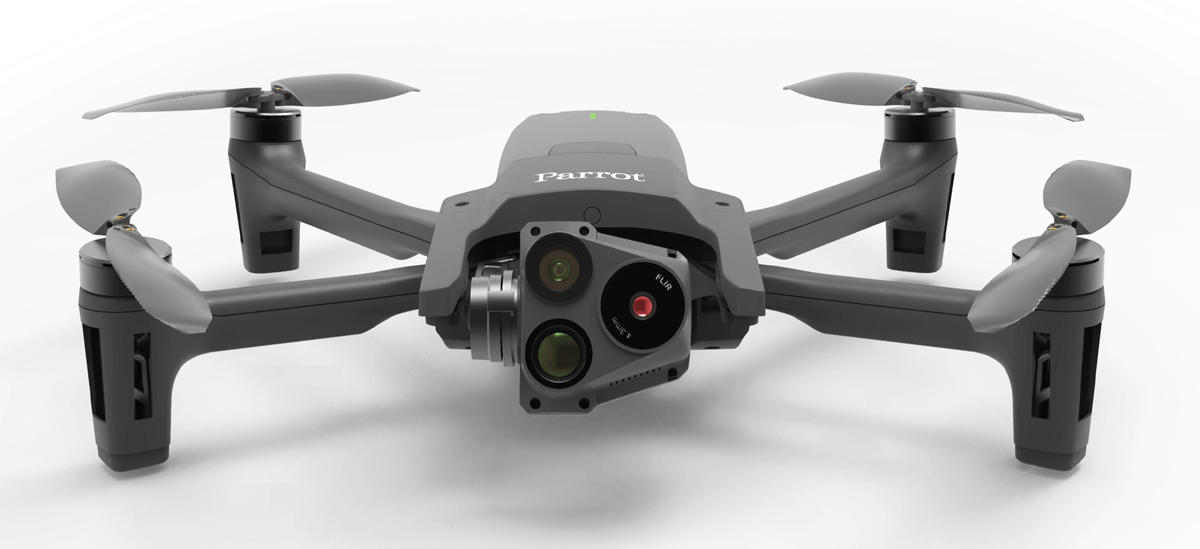}
\endminipage\hfill
\minipage{.14\textwidth}
	\includegraphics[width=\linewidth]{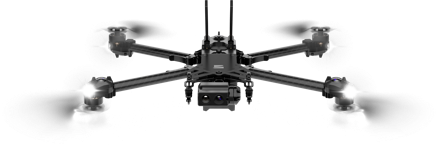}
\endminipage\hfill
\minipage{.14\textwidth}
	\includegraphics[width=\linewidth]{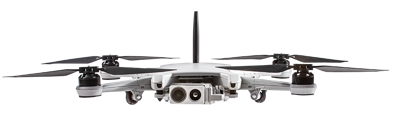}
\endminipage\hfill
\minipage{.14\textwidth}
	\includegraphics[width=\linewidth]{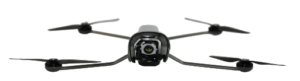}
\endminipage\hfill
\caption{\small{From left to right: Cleo Robotics Dronut, Flyability Elios 2, Lumenier Nighthawk 3, Parrot ANAFI USA GOV, Skydio X2D, Teal Drones Golden Eagle, Vantage Robotics Vesper}}
\label{fig:platforms}
\end{figure}

In our evaluations, as depicted in Fig.~\ref{fig:platforms}, we examine seven small UAS platforms: the Cleo Robotics Dronut~\cite{cleo}, Flyability Elios 2~\cite{elios}, Lumenier Nighthawk 3~\cite{nighthawk}, Parrot ANAFI USA GOV~\cite{parrot}, Skydio X2D~\cite{skydio}, Teal Drones Golden Eagle~\cite{teal}, and Vantage Robotics Vesper~\cite{vantage}. These platforms have been selected due to their variety in capabilities and intended applications. While the Nighthawk 3, Parrot, Skydio X2D, Golden Eagle, and Vesper are intended for outdoor reconnaissance, the  Dronut and Elios 2 are intended for indoor reconnaissance, specifically in urban and industrial environments. In our evaluations, the resulting data has been anonymized by assigning the platforms labels A through G without any specific ordering or correlation. For more information about these platforms, please see references.

\section{Measuring NCAP}
The existing NCAP framework allows us to combine the component and engineering level tests into a predictive measure of UAS autonomous performance. It should be noted that NCAP encapsulates the \emph{potential} of a UAS to operate autonomously, not an evaluation of the UAS's actual task-based autonomous performance. To measure the NCAP score for a UAS platform, we measure its \emph{autonomy level} and combine it with an overall \emph{component performance score}. One of the shortcomings of NCAP is that it simply added the scores to produce a final autonomy score for the system.  
In the following sections, we explain measuring the autonomy level and the component performance denoted as $N_{AL}$ and $N_{CP}$, respectively. Then, we discuss methods for combining these scores and calculating a single autonomy score for our UAS platforms and ranking them accordingly.

\subsection{Measuring NCAP Autonomy Levels ($N_{AL}$)}
The autonomy level, $N_{AL}$, is an overall measure of non-contextual autonomy that helps to classify the system into four classes in range 0 (i.e., no autonomy) to 3 (i.e., full autonomy)~\cite{durst2013development}. NCAP assigns an autonomy level $N_{AL}=0$ to a UAS that only has perception sensors but is not using them to build a model. An example of $N_{AL}=0$ is a UAS equipped with multiple cameras that is operated entirely by teleoperation. A UAS that captures raw data and constructs a model of the task or the environment is assigned an autonomy level of $N_{AL}=1$. An example of $N_{AL}=1$ is a UAS that uses the data from its RGB camera to detect an object in the environment but still requires teleoperation to move through the environment. Given a high-level goal, a UAS with $N_{AL}=2$ uses a planning algorithm and the constructed model to generate a plan. An example of $N_{AL}=2$ is a UAS equipped with a path planning algorithm that enables it to plan paths using a world model but requires a user to select a best path. And a UAS with $N_{AL}=3$ can execute the generated plan without human assistance. A UAS is only considered $N_{AL}=3$, fully autonomous, if it requires no human input during its mission. Based on this classification process, table~\ref{tab:N-AL-table} shows the autonomy level for each UAS in Fig.~\ref{fig:platforms}. It can be seen that the specifications and capabilities of each system in different layers were used to determine the autonomy level for that platform.




\begin{table}[ht]
\scriptsize
\centering
\caption{\small{Evaluation and reasoning of autonomy levels $N_{AL}$ of each UAS platform.}}
\label{tab:N-AL-table}
\begin{tabular}{cllllc}
\toprule
\textbf{Platform} & \textbf{Perception} & \textbf{Modeling} & \textbf{Planning} & \textbf{Execution} & $N_{AL}$ \\ \midrule
UAS A             & \begin{tabular}[c]{@{}l@{}}2 RGB camera, thermal \\ camera, LiDAR, GPS\end{tabular}                                  & SLAM capabilities                                                                       & \begin{tabular}[c]{@{}l@{}} Obstacle avoidance \\ planning\end{tabular}      & \begin{tabular}[c]{@{}l@{}}Avoids obstacles, \\ auto-land\end{tabular}            & 3                       \\ \hline
UAS B             & \begin{tabular}[c]{@{}l@{}}RGB camera, IMU, \\ thermal camera, \\ 5 distance sensors\end{tabular} & \begin{tabular}[c]{@{}l@{}}Models surroundings \\ using distance sensors\end{tabular}   & None                                                                            & None                                                                            & 1                       \\ \hline
UAS C             & \begin{tabular}[c]{@{}l@{}} RGB camera, thermal \\ camera  \end{tabular}                                                                                & None                                                                                    & None                                                                            & None                                                                              & 0                       \\ \hline
UAS D             & \begin{tabular}[c]{@{}l@{}}2 RGB cameras, IMU \\ thermal camera, GPS\end{tabular}                   & \begin{tabular}[c]{@{}l@{}}Uses GPS and IMU \\ for positioning in maps, \\visual modeling of targets\end{tabular} & \begin{tabular}[c]{@{}l@{}}Geofencing and \\ autonomous navigation\end{tabular} & \begin{tabular}[c]{@{}l@{}}Target tracking, \\ return to home\end{tabular} & 3                       \\ \hline
UAS E             & \begin{tabular}[c]{@{}l@{}}6 RGB cameras, thermal \\ camera, GPS  \end{tabular}   & \begin{tabular}[c]{@{}l@{}}Maps surrounding areas \\ from camera imaging\end{tabular}   & \begin{tabular}[c]{@{}l@{}}Plans best path \\ in environment\end{tabular}       & \begin{tabular}[c]{@{}l@{}}Obstacle avoidance, \\ path execution\end{tabular}     & 3                       \\ \hline
UAS F             & \begin{tabular}[c]{@{}l@{}} RGB camera, thermal \\camera, GPS, IMU   \end{tabular}                                                                               & None & None                                                                            & None                                                                              & 0                       \\ \hline
UAS G             & \begin{tabular}[c]{@{}l@{}} RGB camera, thermal \\ camera, GPS  \end{tabular}                                                                                      & None & None                                                                            & None                                                                              & 0                       \\ \hline
\end{tabular}
\end{table}

\subsection{Measuring NCAP Component Performance ($N_{CP}$)}
To calculate the $N_{CP}$, we have selected a set of important features including flight time, charging time, RGB streaming resolution, Field of View (FOV), maximum range, thermal camera resolution, weight, maximum flight speed, number of sensors, and number of pre-programmed (i.e. \emph{smart}) behaviors. Table~\ref{tab:ncap-measurements-table} shows the  extracted data for the set of features for each UAS platform. It has to be noted that any number or combination of features can be included in the calculation of $N_{CP}$ and the calculation is not limited to the specific set of features selected here. 
Various methods can be used for the aggregation of the data in Table~\ref{tab:ncap-measurements-table} and since NCAP does not specify a particular combination method, we investigate five methods in the next section. 

\vspace{-1em}
\begin{table}[ht]
\setlength{\tabcolsep}{4pt} 
\centering
\scriptsize
\caption{\small{Selected UAS platform features used for $N_{CP}$  measurements}}
\label{tab:ncap-measurements-table}
\begin{tabular}{lcccccccccc}
\toprule
\multicolumn{1}{c}{\textbf{Platform}} &
\textbf{\begin{tabular}[c]{@{}c@{}}Flight \\ Time \\ (min)\end{tabular}} &
\textbf{\begin{tabular}[c]{@{}c@{}}Charge \\ Time \\ (min)\end{tabular}} & 
\textbf{\begin{tabular}[c]{@{}c@{}}Stream \\  Res.\end{tabular}} & 
\textbf{\begin{tabular}[c]{@{}c@{}}FOV \end{tabular}} & 
\textbf{\begin{tabular}[c]{@{}c@{}}Max. \\ Range \\ (m) \end{tabular}} &
\textbf{\begin{tabular}[c]{@{}c@{}}Thermal \\ Camera \\ Res.\end{tabular}} &
\textbf{\begin{tabular}[c]{@{}c@{}}Weight \\ (g) \end{tabular}} &
\textbf{\begin{tabular}[c]{@{}c@{}}Max Flight \\ Speed \\ (m/s) \end{tabular}} &
\textbf{\begin{tabular}[c]{@{}c@{}} \# of \\ Sensors \end{tabular}} &
\textbf{\begin{tabular}[c]{@{}c@{}} \# of \\ Smart \\ Behaviors \end{tabular}} \\ 
\toprule

UAS A & 15 & 50 & FHD & 100\degree & 2000 & N/A & 370 & 3 & 3 & 2  \\ 
UAS B & 10 & 90 & FHD30p & 114\degree & 500 & 160$\times$120 & 1450 & 6.5 & 10 & 7  \\ 
UAS C & 22 & -  & FHD & - & 2000 & 620x512 & 1200  & 3 & 4 & 5  \\ 
UAS D & 32 & 120 & HD & 84\degree & 4000 & 320$\times$256 & 500 & 14.7 & 10 & 5   \\ 
UAS E & 23 & 120  & 4k60p & 200\degree & 3500 & 320p & 775 & 16 & 11 & 10   \\ 
UAS F & 30 & 45 & 4k & 90\degree & 3000 & 320$\times$256 & 1044 & 22 & 7 & 3  \\ 
UAS G & 50 & -  & FHD & 63\degree & 4000 & 320p & 697 & 20 & 6 & 2   \\ 
\bottomrule
\end{tabular}
\end{table}
\vspace{-1em}

\subsection{Methods for Combining Scores}
\label{sec:combin_method}
In this section, we discuss methods for combining test scores and apply them to the obtained data in Table~\ref{tab:ncap-measurements-table} for the calculation of $N_{CP}$. The most common method is to normalize the data and use a weighted sum~\cite{tofallis2014add}. However, one of the main disadvantages of the weighted normalized sum is that different normalization techniques sometimes result in dissimilar scores. Additionally, we show that calculating the autonomy score using a weighted product results is more favorable results.


\subsubsection{Weighted Normalized Sum}
Weighted Normalized Sum is the most common method used for combining scores for a set of features denoted by $\phi_i$ for $i \in [0,N]$ where $N$ is the number of features. The normalization step is required to make features in different units comparable. Several normalization techniques exist. We employ and compare four techniques: 
\begin{enumerate}[topsep=0pt,itemsep=0pt,parsep=0pt,partopsep=0pt]
    \item Divide by maximum ($\eta_{max}$). This normalization technique converts the maximum value to $1$ and the rest of values become a number less than $1$. 
    \item Divide by the sum ($\eta_{sum}$). This normalization technique produces a proportional value for each original number.
    \item Range mapping from the current to $[0, 1]$ ($\eta_{map}$). This normalization technique considers both the minimum and maximum values and maps them to $0$ and $1$, respectively. All other values are converted to a number in range $[0, 1]$.
    \item z-score ($\eta_{zsc}$). In this statistical normalization technique, also known as standard score, we first subtract the mean value from the data and then divide all values by the standard deviation of the data. 
\end{enumerate}

The weighted normalized sum can be represented as $S_k = \sum_i^{N} w_i \eta_{k}(\phi_i)$ where $k$ represents one of the normalization techniques ($k \in \{ \textrm{max, sum, map, zsc} \}$). In our experiments, we use two weighting schemes: (a) uniform where where each feature is assigned a weight value of $w_i=1/N$, and (b) a user-defined  vector $W$ based on preferences and importance of each feature as follows: $W = [0.07, 0.03, 0.10, 0.10, 0.05, 0.10, 0.05, 0.05, 0.15, 0.30]$, with indices corresponding to feature indices in Table~\ref{tab:ncap-measurements-table}. 

It has to be noted that increasing and decreasing the value of the features can have reverse effect of the autonomy of the system. For the features that do not comply with the ``more is better'' rule (e.g., charging time), we negated the effect of that feature in the sum. This could be interpreted as a negative weight value too. The resulting evaluations of the component performance $N_{CP}$ for each normalization technique using the uniform weights and the user-defined weights can be seen in Table~\ref{tab:ncp-score-table-equal} and Table~\ref{tab:ncp-score-table-norm}, respectively. These tables also show that platform rankings in parenthesis and it can be seen that different normalization techniques result is different rankings within the same data. To give a perspective the $N_{AL}$ scores from Table~\ref{tab:N-AL-table} were included in both tables.

\begin{table}[b]
\renewcommand\arraystretch{1.0}
\setlength{\tabcolsep}{1.5pt} 
\begin{minipage}{.48\linewidth}
\centering
\scriptsize
\caption{\small{$N_{CP}$ score of each UAS using different combination methods and uniform weights.}}
\label{tab:ncp-score-table-equal}
\begin{tabular}{lcccccc}
\toprule
\multicolumn{1}{c}{\textbf{Platform}} &
$S_{max}$ & $S_{map}$ & $S_{zsc}$ & $S_{sum}$ & $P$ & $N_{AL}$ \\ \midrule
UAS A & 0.18 (6) & 0.33 (5) & -0.98 (7) & 0.05 (7) & 2.48 (6) & 3 \\ 
UAS B & 0.17 (7) & 0.30 (7) &  0.22 (2) & 0.05 (6) & 2.29 (7) & 1 \\ 
UAS C & 0.20 (5) & 0.30 (6) & -0.09 (5) & 0.06 (5) & 2.60 (5) & 0 \\
UAS D & 0.35 (4) & 0.50 (4) &  0.05 (4) & 0.09 (4) & 3.48 (4) & 3 \\ 
UAS E & 0.53 (1) & 0.72 (1) &  0.93 (1) & 0.14 (1) & 4.63 (1) & 3 \\
UAS F & 0.39 (2) & 0.57 (2) &  0.05 (3) & 0.10 (3) & 3.81 (2) & 0 \\
UAS G & 0.38 (3) & 0.52 (3) & -0.27 (6) & 0.10 (2) & 3.56 (3) & 0 \\
\bottomrule
\end{tabular}
\end{minipage}
\begin{minipage}{.48\linewidth}
\centering
\scriptsize
\caption{\small{$N_{CP}$ score of each UAS using different combination methods and user-defined weights.}}
\label{tab:ncp-score-table-norm}
\begin{tabular}{lcccccc}
\toprule
\multicolumn{1}{c}{\textbf{Platform}} &
$S_{max}$ & $S_{map}$ & $S_{zsc}$ & $S_{sum}$ & $P$ & $N_{AL}$ \\ \midrule
UAS A & 0.24 (7) & 0.19 (7) & -0.98 (7) & 0.06 (7) & 3.17 (7) & 3\\ 
UAS B & 0.43 (4) & 0.43 (3) & -0.08 (4) & 0.12 (3) & 4.66 (4) & 1\\ 
UAS C & 0.39 (6) & 0.34 (6) & -0.14 (6) & 0.11 (6) & 4.51 (5) & 0\\ 
UAS D & 0.48 (2) & 0.47 (2) &  0.01 (3) & 0.13 (2) & 5.36 (2) & 3\\ 
UAS E & 0.78 (1) & 0.86 (1) &  1.28 (1) & 0.21 (1) & 8.51 (1) & 3	\\ 
UAS F & 0.44 (3) & 0.43 (4) & -0.10 (5) & 0.11 (4) & 5.01 (3) & 0	\\ 
UAS G & 0.42 (5) & 0.38 (5) &  0.16 (2) & 0.11 (5) & 4.39 (6) & 0\\
\bottomrule
\end{tabular}
\end{minipage}
\end{table}

\subsubsection{Weighted Product}
Each of these normalization methods have known disadvantages~\cite{tofallis2014add}. For instance, unlike the divide by sum technique, neither the range mapping nor the z-score preserve proportionality of the data. z-score is also extremely sensitive to outlier data points. Consequently, we investigate the weighted product which is an alternative method used for combining scores. The Weighted product is represented as $P=\prod_i^N \phi_i^{w_i}$ and does not require a normalization step because rescaling has no effect on the outcome. Similar to the weighted normalized sum method, negative weights represent the ``less is better'' features. Unlike the weighted normalized sum methods, in the weighted product method the weights do not depend on the units of measurement of the features~\cite{tofallis2014add}. The UAS platform $N_{CP}$ scores and their corresponding rankings using the weighted product method have been also reported in Table~\ref{tab:ncp-score-table-equal} and Table~\ref{tab:ncp-score-table-norm}, compared alongside summing methods. It can be seen that all methods agree on UAS E being rank $1$ but there is no consensus on any other UAS platform's ranking.


\subsection{Combined NCAP Score vs. NCAP Coordinate}
\begin{figure}[t]
\centering
\includegraphics[width=0.48\textwidth]{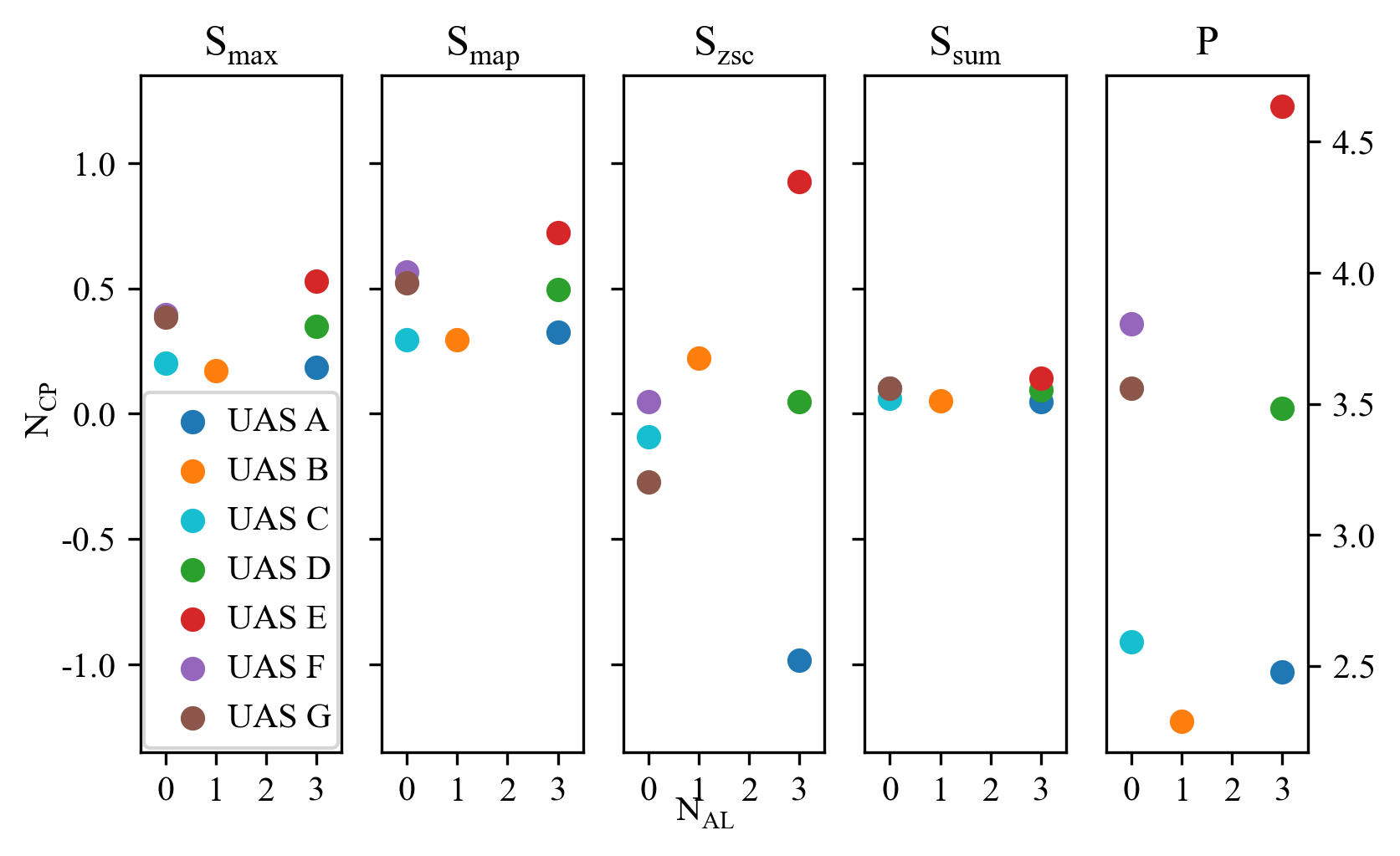}
\includegraphics[width=0.48\textwidth]{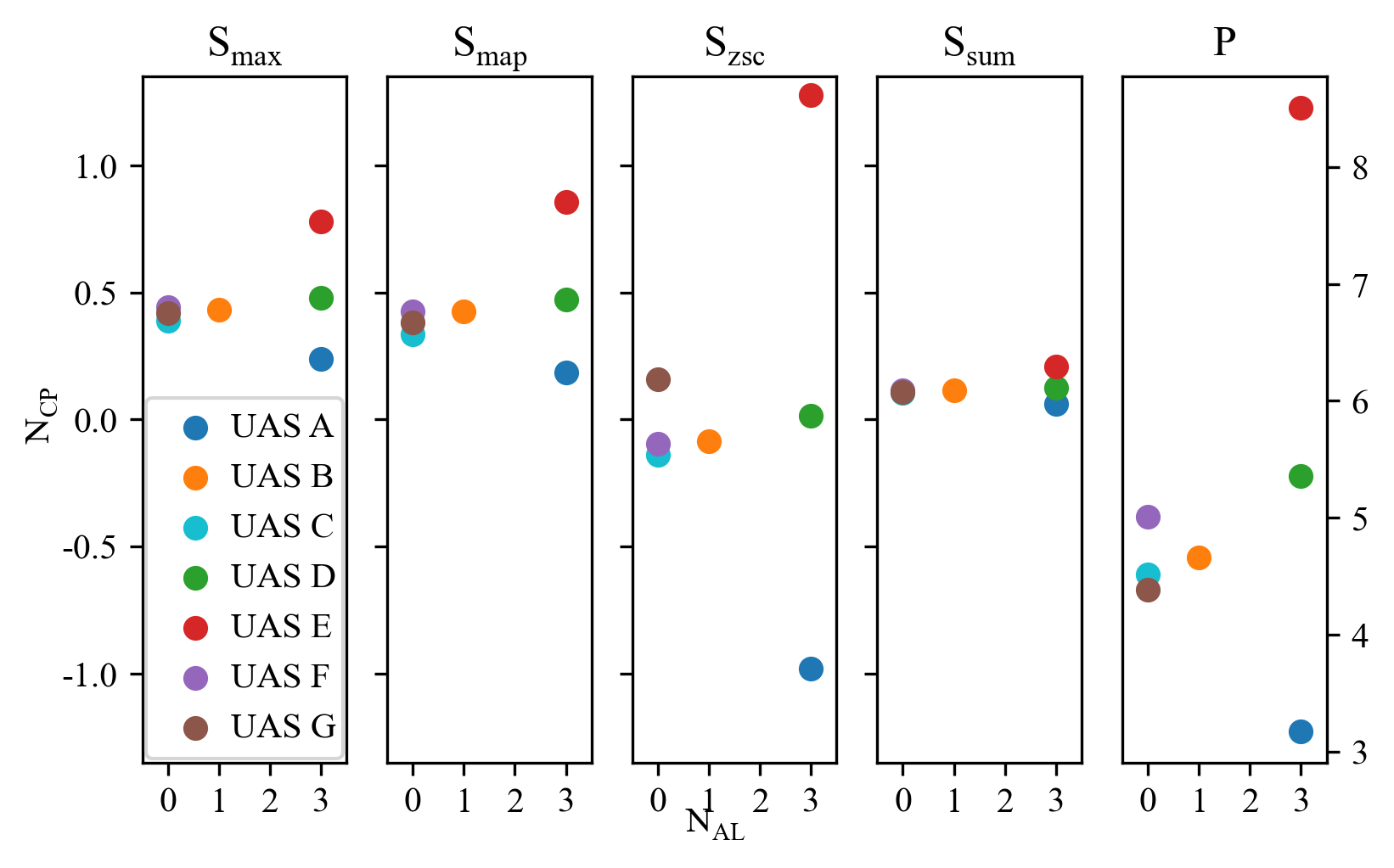}
\caption{\small{UAS platform autonomy measure represented in the NCAP coordinate $<N_{AL}, N_{CP}>$ with uniform weight values (left) and user-defined weight values (right).}}
\label{fig:autonomy_plot}
\end{figure}

The combined NCAP score is calculated by combining the obtained autonomy level, $N_{AL}$, with the component performance score, $N_{CP}$. The combination of these scores allows for a final score which comprehensively incorporates the system's overall potential autonomy level  with the potential of its components. Originally, NCAP overall autonomy score is obtained by simply adding the two evaluations $N_{AL}$ and $N_{CP}$~\cite{durst2014levels}. One of the issues with this method is that $N_{AL}$ and $N_{CP}$ represent very different measures. $N_{AL}$ classifies platforms into different classes and partially represents a qualitative evaluation of the systems, however, $N_{CP}$ provides a quantitative evaluation of the system components. To address this problem, we represent the modified NCAP score in an \emph{NCAP coordinate} with $N_{AL}$ as the $x$-axis and $N_{CP}$ as the $y$-axis. Fig.~\ref{fig:autonomy_plot} illustrates the NCAP coordinate for each UAS using the discussed combining methods and two feature weight vectors.


\begin{table}[b]
\renewcommand\arraystretch{1.0}
\setlength{\tabcolsep}{4pt} 
\begin{minipage}{.48\linewidth}
\centering
\scriptsize
\caption{\small{Relative potential autonomy distance with respect to UAS E for uniform weights.}}
\label{tab:auton_diff}
\begin{tabular}{lccccc}
\toprule
\multicolumn{1}{c}{\textbf{Platform}} &
$AD_{max}$ & $AD_{map}$ & $AD_{zsc}$ & $AD_{sum}$ & $AD_{P}$ \\ \midrule
UAS A & 0.34 & 0.40 & 1.91 & 0.09 & 2.16\\ 
UAS B & 2.03 & 2.05 & 2.12 & 2.00 & 3.08\\  
UAS C & 3.02 & 3.03 & 3.17 & 3.00 & 3.63\\ 
UAS D & 0.18 & 0.23 & 0.88 & 0.05 & 1.15\\ 
UAS E & 0    & 0    & 0    & 0    & 0   \\ 
UAS F & 3.00 & 3.00 & 3.13 & 3.00 & 3.11\\ 
UAS G & 3.01 & 3.01 & 3.23 & 3.00 & 3.19\\ 
\bottomrule
\end{tabular}
\end{minipage}
\begin{minipage}{.48\linewidth}
\centering
\scriptsize
\caption{\small{Relative potential autonomy distance with respect to UAS E for user-defined weights.}}
\label{tab:auton_diff-user}
\begin{tabular}{lccccc}
\toprule
\multicolumn{1}{c}{\textbf{Platform}} &
$AD_{max}$ & $AD_{map}$ & $AD_{zsc}$ & $AD_{sum}$ & $AD_{P}$ \\ \midrule
UAS A & 0.54 & 0.67 & 2.26 & 0.15 & 5.34\\ 
UAS B & 2.03 & 2.05 & 2.42 & 2.00 & 4.34\\  
UAS C & 3.03 & 3.04 & 3.32 & 3.00 & 5.00\\ 
UAS D & 0.30 & 0.38 & 1.26 & 0.08 & 3.15\\ 
UAS E & 0    & 0    & 0    & 0    & 0   \\ 
UAS F & 3.02 & 3.03 & 3.30 & 3.00 & 4.61\\ 
UAS G & 3.02 & 3.04 & 3.20 & 3.00 & 5.10\\ 
\bottomrule
\end{tabular}
\end{minipage}
\end{table}

\subsection{Potential Autonomy Distance}
Representing the autonomy scores in the NCAP space allows for assigning an absolute measure of a system's overall autonomy that we call the \emph{Potential Autonomy Distance} (AD) and measure as the Euclidean distance from a system's NCAP coordinate, $<N_{AL},N_{CP}>$, to the origin, $<0,0>$. Since both $N_{AL}$ and $N_{CP}$ are non negative values, the bigger autonomy distance represent a system with higher potential autonomy. 

Additionally, this representation allows for determining a relative measure of system autonomy when comparing multiple systems. To obtain this measure, we first calculate the potential autonomy distance (AD) for all the platforms, then select the platform with the highest distance as the reference, and find the relative AD differences between the reference and of all other platforms. In our experiments, it can be seen that the UAS E has the highest absolute autonomy distance, so we selected it as our reference. The resulting relative autonomy measure can be seen in Tables~\ref{tab:auton_diff} and \ref{tab:auton_diff-user}. A higher relative AD measure indicates more distance between that system and the system with the highest autonomy score and less distance to non-autonomy level (i.e., $<0,0>$).


\section{Conclusions}
We have investigated five methods of combining features,  and found that the weighted product method provides a consistent combination of the values. We also introduced the non-contextual autonomy coordinate and represented the overall autonomy of a system using an autonomy distance. We applied our method to a set of seven UAS and obtained their absolute autonomy scores as well as their relative scores with respect to the best system. 
\subsubsection*{Acknowledgement}
This work is sponsored by the Department of the Army, U.S. Army Combat Capabilities Development Command Soldier Center, award number W911QY-18-2-0006.

\bibliographystyle{plain}
\bibliography{references}

\end{document}